\pdfoutput=1
\documentclass[11pt]{article}

\usepackage[preprint]{acl}

\usepackage{times}
\usepackage{latexsym}

\usepackage[T1]{fontenc}
\usepackage[utf8]{inputenc}

\usepackage{microtype}

\usepackage{inconsolata}

\usepackage{graphicx}

\usepackage{enumitem}
\usepackage{amsmath}
\usepackage{xspace}
\usepackage{hyperref}
\usepackage{multirow}
\usepackage{booktabs}
\usepackage{array}
\usepackage[normalem]{ulem}
\useunder{\uline}{\ul}{}
\usepackage{xcolor}
\usepackage{tabularx}

\newcommand{\Data}{\mathcal{D}\xspace}
\newcommand{\ParaData}{\mathcal{D}_{\text{para}}\xspace}
\newcommand{\TrfData}{\mathcal{D}_{\text{trf}}\xspace}

\newcommand{\SFTModel}{f_{\text{SFT}}\xspace}
\newcommand{\ParaModel}{f_{\text{para}}\xspace}
\newcommand{\SFTModelTilde}{f^{\rightarrow s}_{\text{inv}}\xspace}
\newcommand{\RefModel}[1]{f^{#1}_{\text{ref}}\xspace}
\newcommand{\POModel}[1]{f^{#1}_{\text{PO}}\xspace}
\newcommand{\POData}[1]{\mathcal{D}^{#1}_{\text{PO}}\xspace}

\title{Style Transfer with Multi-iteration Preference Optimization}

\author{Shuai Liu \and Jonathan May \\
        Information Sciences Institute \\ University of Southern California \\ \texttt{\{liushuai, jonmay\}@isi.edu}}

\begin{document}
\maketitle

\begin{abstract}
Numerous recent techniques for text style transfer characterize their approaches as variants of reinforcement learning and preference optimization.
In this work, we consider the relationship between these approaches and a class of optimization approaches developed primarily for (non-neural) statistical machine translation, formerly known as `tuning'.
Inspired by these techniques from the past, we improve upon established preference optimization approaches, incorporating multiple iterations of exploration and optimization, and choosing contrastive examples by following a `hope' vs `fear' sampling strategy. 
Cognizant of the difference between machine translation and style transfer, however, we further tailor our framework with a new pseudo-parallel generation method and a dynamic weighted reward aggregation method to tackle the lack of parallel data and the need for a multi-objective reward.
We evaluate our model on two commonly used text style transfer datasets. 
Through automatic and human evaluation results we show the effectiveness and the superiority of our model compared to state-of-the-art baselines.
\end{abstract}
\begin{figure*}[t!]
    \centering
    \includegraphics[width=2.0\columnwidth]{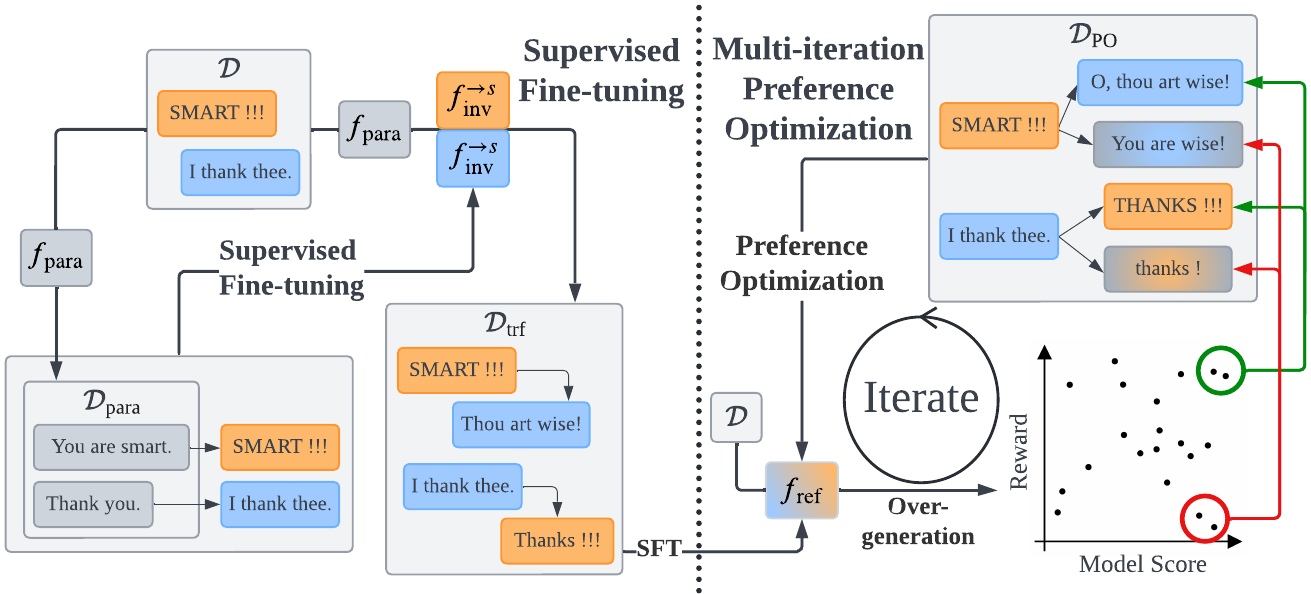}
    \caption{
        An overview of STAMP, in which we first train a unified style transfer model using supervised fine-tuning on pseudo-parallel data generated from non-parallel data, and then further train the model using multi-iteration preference optimization on preference pairs constructed with hope-and-fear sampling.
    }
    \label{fig:stamp_overview}
    \vspace{-0.4cm}
\end{figure*}

\section{Introduction}

Text style transfer aims to rewrite a given text to match a specific target style while preserving the original meaning.
This task has drawn significant attention recently due to its broad range of applications, such as text simplification \cite{laban-etal-2021-keep}, formality transfer \cite{rao-tetreault-2018-dear, liu-etal-2022-semi}, text detoxification \cite{dale-etal-2021-text, hallinan-etal-2023-detoxifying}, authorship transfer \cite{patel2023lowresource, liu2024authorship}, and authorship anonymization \cite{217499, bo-etal-2021-er}.
%Despite numerous approaches having been developed,
Recent approaches have focused on pseudo-parallel data generation \cite{krishna-etal-2020-reformulating, riley-etal-2021-textsettr} and policy optimization \cite{gong-etal-2019-reinforcement, liu-etal-2021-learning}.
STEER \cite{hallinan-etal-2023-steer} and ASTRAPOP \cite{liu2024authorship} combine the two and achieve state-of-the-art performance on text style transfer and authorship style transfer, respectively.

In this work, we seek to advance the frontier of text style transfer, drawing inspiration from the optimization techniques developed in the era of statistical phrasal machine translation, in which the lack of correlation between the log-linear model objective and the desired evaluation metric, typically BLEU \cite{papineni-etal-2002-bleu}, was observed \cite{och-2003-minimum}.
Approaches to align\footnote{not to be confused with word alignment.} the two objectives came to be known as \textit{tuning},\footnote{not to be confused with parameter fine-tuning.} beginning with \citet{och-2003-minimum}, and evolving into  online variants \citep{chiang-etal-2008-online}, rank-based approaches \citep{hopkins-may-2011-tuning}, batch-based approaches \cite{cherry-foster-2012-batch}, and several others.
Tuning methods follow a generate-and-optimize pattern: a model is used to generate multiple candidate hypotheses per input, and then parameters are adjusted such that the argmax according to the model score also maximizes the evaluation metric.
In this regard, tuning methods resemble approaches taken in the application of policy optimization algorithms, such as PPO \cite{schulman2017proximal}, to generative language modeling \cite{ouyang2022training}.
More recent algorithms, such as DPO \cite{rafailov2023direct} and CPO \cite{xu2024contrastive}, which replace reinforcement learning (RL) in PPO with \textit{preference} optimization (PO), 
%\footnote{which refers exclusively to preference optimization and never to policy optimization.} 
are reminiscent of the pairwise ranking optimization approach to tuning \cite{hopkins-may-2011-tuning}.
Given this close relationship between these approaches, we can consider whether other techniques developed to improve MT tuning could be applied to optimization for style transfer. 

In this work, we propose Style TrAnsfer with Multi-iteration Preference optimization (STAMP), a two-phase PO training framework, in which we first use supervised fine-tuning to build a reference model from pseudo-parallel data and then train the reference model using PO.
STAMP is similar to STEER and ASTRAPOP at a high level but is enhanced with two techniques borrowed from MT tuning and two modifications that further adapt it for text style transfer.
First, we include \textit{multiple iterations} of preference pair generation followed by model optimization \cite{och-2003-minimum}, which has already been shown to be effective on other Seq2Seq tasks such as mathematical and scientific reasoning \cite{chen2024selfplay, pang2024iterative, song2024trial, yuan2024selfrewarding}.
Second, following the hope-and-fear sampling in \citet{JMLR:v13:chiang12a}, for PO, we over-generate outputs using the reference model and construct preference pairs using samples with high model scores and extreme (high or low) task objective scores, in order to avoid dangerous generation and encourage reachable good generation.
To improve the quality of the reference model and the balance across the multiple training objectives, we additionally design a new two-step end-to-end pseudo-parallel data generation method and a dynamic reward aggregation method.

We evaluate our model on two popular text style transfer datasets, Grammarly’s Yahoo Answers Formality Corpus (GYAFC) \cite{rao-tetreault-2018-dear} and the Corpus of Diverse Styles (CDS) \cite{krishna-etal-2020-reformulating}.
Extensive experiments show that our model performs well on both in-domain and out-of-domain text style transfer, and outperforms all state-of-the-art baselines on both datasets.

%\vspace{5pt}

Our main contributions are:
\begin{itemize}[itemsep=3pt, topsep=3pt, parsep=0pt]
  \item We propose a multi-iteration contrastive preference optimization training framework with hope-and-fear preference pair construction for text style transfer.
  \item We design a new pseudo-parallel generation strategy and a dynamic weighted rewarded aggregation method to enhance the training framework for text style transfer.
  \item We show that, with the enhancements, our training framework produces style transfer models that achieve state-of-the-art performance on two popular text style transfer datasets.\footnote{Code and models sufficient for a reproducibility study are available at \url{https://github.com/isi-nlp/STAMP}.}
\end{itemize}

\section{Methodology}

In this section, we formalize the text style transfer task and introduce our training framework, STAMP.

\subsection{Task Definition}
Given a source text $\textbf{x}$ and a desired target style $s$, the goal of text style transfer is to generate a fluent rewrite of $\textbf{x}$, denoted as $\textbf{x}^{\rightarrow s}$, that has the same meaning as $\textbf{x}$ but is in style $s$.
In this work, we focus on high-resource text style transfer in which we have access to a reasonable number of texts\footnote{In this work, we assume at least 2000 texts per style.} for each target style.
Specifically, we have a set of texts with style labels, denoted as $\Data=\{(\textbf{x}_1, s_1), \cdots, (\textbf{x}_n, s_n)\}$, where $\textbf{x}_i$ and $s_i$ refer to the $i^{\text{th}}$ text and its style, respectively.
For convenience, we adopt notations from \citet{hallinan-etal-2023-steer} and denote the \textbf{fluency} of a text $\textbf{x}_i$ as $\text{F}(\textbf{x}_i)$, the \textbf{meaning similarity} between two texts $\textbf{x}_i$ and $\textbf{x}_j$ as $\text{MS}(\textbf{x}_i, \textbf{x}_j)$, and the \textbf{target style strength} of a text $\textbf{x}_i$ w.r.t.\  a target style $s$ as $\text{TSS}(\textbf{x}_i, s)$.
Thus, given  $\Data$, we aim to build a text style transfer system that maximizes three independent objectives: $\text{F}(\textbf{x}^{\rightarrow s})$, $\text{MS}(\textbf{x}, \textbf{x}^{\rightarrow s})$, and $\text{TSS}(\textbf{x}^{\rightarrow s}, s)$.\footnote{For brevity, we omit the arguments where unambiguous.}

\subsection{Framework Overview}
STAMP is a preference optimization-based training framework that contains two main stages, a supervised fine-tuning (SFT) stage and a multi-iteration preference optimization (PO) stage.
In the SFT stage, we first generate a dataset $\TrfData$ of end-to-end pseudo-parallel style transfer pairs from the (non-parallel) dataset $\Data$ and then train a style transfer model $\SFTModel$ on $\TrfData$ using supervised fine-tuning.
In the PO stage, we train a model initialized to $\SFTModel$ using multi-iteration PO\footnote{See \autoref{sec:implementation} for details on the choice of PO used here.} to directly maximize the three objectives, TSS, MS, and F, and obtain our final transfer model $\POModel{}$.

\subsection{Supervised Fine-tuning}
\label{sec:sft}
Due to a lack of parallel data, we adopt the technique described by \citet{krishna-etal-2020-reformulating}, in which style-oriented paraphrasing is used to generate pseudo-parallel transfer data for each target style.
Specifically, we paraphrase the texts in $\Data$ using a general paraphraser $\ParaModel$ similar to \citet{krishna-etal-2020-reformulating} and \citet{hallinan-etal-2023-steer}.
To ensure meaning similarity preservation of the paraphrases, we generate $k_{\text{para}}$ paraphrases for each text $\textbf{x}_i \in \Data$ and select the one with the highest meaning similarity to the original text, denoting it $\textbf{p}_i$.
We then obtain a dataset of paraphrases $\ParaData=\{\textbf{p}_1, \cdots, \textbf{p}_n\}$.
For each target style $s$, we train a Seq2Seq model $\SFTModelTilde$\footnote{`inverse' due to data provenance, c.f.\ \cite{krishna-etal-2020-reformulating}} 
on $\{(\textbf{p}_i\rightarrow \textbf{x}_i) \mid 0 \leq i \leq n \text{ and } s_i=s\}$ to maximize
\begin{equation}
    p(\textbf{x} \mid \textbf{p}) = \prod^{\left|\textbf{x}\right|}_{i=1}{p(\textbf{x}[i] \mid \textbf{p},\textbf{x}[<\!i])}
\end{equation}
where $\textbf{x}[i]$ and $\textbf{x}[<\!i]$ represent the $i^{\text{th}}$ token in \textbf{x} and tokens preceding the $i^{\text{th}}$ token in \textbf{x}, respectively. 

Following \citet{krishna-etal-2020-reformulating}, we can transfer the style of a text $\textbf{x}$ to a style $s$ through
\begin{equation}
   \textbf{x}^{\rightarrow s} = \SFTModelTilde(\ParaModel(\textbf{x})) 
   \label{eq:2-step_transfer}
\end{equation}
where $\textbf{x}^{\rightarrow s}$ is the transferred text.
However, the two-step generation breaks the gradient connection between $\textbf{x}$ and $\textbf{x}^{\rightarrow s}$ which is needed in the PO stage to maximize the meaning similarity between $\textbf{x}$ and $\textbf{x}^{\rightarrow s}$.
Therefore, we need an end-to-end pseudo-parallel dataset $\TrfData$ to train a model that directly transfers a source text to each target style with no intermediate step.

To obtain $\TrfData$, we transfer the texts in $\Data$ using $\ParaModel$ and $\SFTModelTilde$ for each target style $s$.
Specifically, for each target style $s$, we transfer the texts in other styles in $\Data$ using \autoref{eq:2-step_transfer} and obtain a dataset of style transfer pairs $\TrfData^{\rightarrow s}=\{(\textbf{x}_i\rightarrow \textbf{t}_i, s) \mid (\textbf{x}_i, s_i)\in\Data \; \text{and} \; s_i\neq s\}$, where $\textbf{t}_i = \SFTModelTilde(\ParaModel(\textbf{x}_i))$ is a transfer of $x_i$ in style $s$.
To obtain high-quality transferred texts, we generate $k_{\text{sft}}$ transfers for each source text and select the one with the highest $\text{F}\cdot \text{MS}^{\tau_{\text{ms}}}\cdot \text{TSS}$, where  $\tau_{\text{ms}} > 1$ is a temperature hyperparameter incorporated into the MS term to emphasize meaning similarity.
We then construct $\TrfData$ by combining $\TrfData^{\rightarrow s}$ for all target styles and train an end-to-end style transfer model $\SFTModel$ on the combined data $\TrfData$ to maximize
\begin{equation}
    p(\textbf{t} \mid \textbf{x}) = \prod^{\left|\textbf{t}\right|}_{i=1}{p(\textbf{t}[i] \mid \textbf{x},\textbf{t}[<\!i], s)}
    \label{eq:1-step_transfer}
\end{equation}
Note that unlike \autoref{eq:2-step_transfer}, the probability in \autoref{eq:1-step_transfer} is also conditioned on $s$ because we adopt the unified model setting in \cite{hallinan-etal-2023-steer}.
That is, we have a single transfer model for all target styles and control the target style with control codes.

\subsection{Multi-iteration Preference Optimization}
We further train the SFT model $\SFTModel$ from the previous stage with multi-iteration PO to directly optimize the model on the style transfer objectives: F, MS, and TSS. 
To apply PO \cite{rafailov2023direct,xu2024contrastive} we first generate paired preference data from a \textit{reference model} $\RefModel{}$ and then train a model on this offline preference data in a contrastive manner starting from the reference model.
Inspired by \citet{och-2003-minimum} and recent studies in iterative PO, such as \citet{yuan2024selfrewarding} and \citet{chen2024selfplay}, we perform PO for multiple iterations to improve over the offline-only training, updating the reference model between iterations. Specifically, in iteration $i$, we construct preference dataset $\POData{i}$ by transferring texts drawn from $\Data$, using reference model $\RefModel{i}$. We use PO \cite{rafailov2023direct,xu2024contrastive} to train a model initialized to $\RefModel{i}$ to match the preferences in $\POData{i}$; we call the resulting model $\POModel{i}$. We define $\RefModel{1}$ to be $\SFTModel$ and in all other  cases we define $\RefModel{i}$ to be $\POModel{i-1}$. We next detail how the preference pairs in $\POData{i}$ are constructed and the reward function used in this process.

\subsubsection{PO Data Generation}
\label{sec:pogen}
We construct the preference dataset from $\Data$ using the hope-and-fear sampling strategy in \citet{JMLR:v13:chiang12a}. While that work used BLEU \cite{papineni-etal-2002-bleu} as a preference metric, we instead use our style transfer reward $\mathcal{R}$ which is detailed in \autoref{sec:reward}.
Specifically, for each style $s$, we generate $k_{\text{PO}}$ rewrites of each text $\textbf{x}_i$ in $\Data$, whose initial style $s_i \neq s$, into style $s$ and select the preference pair from the rewrites based on both the reward scores $\mathcal{R}$ and the model scores $\mathcal{M}$ of the rewrites, where $\mathcal{M}$ is the average token-level probability w.r.t.\ $\RefModel{}$.
We select the rewrite with the highest $\mathcal{M}^{\tau_{\mathcal{M}}} + \mathcal{R}$ as the ``winning'' rewrite $\textbf{t}^{w}_i$ and the rewrite with the highest $\mathcal{M}^{\tau_{\mathcal{M}}} - \mathcal{R}$ as the ``losing'' rewrite\footnote{also called ``chosen'' and ``rejected'' rewrites in PO literature \cite[e.g.,][]{rafailov2023direct}.} $\textbf{t}^{l}_i$, where $\tau_{\mathcal{M}}$ is the temperature controlling the weight of model score.\footnote{In practice, we find using model score does not benefit performance, so we drop this term for STAMP, which reduces the preference pair selection criteria to the sample with the highest $\mathcal{R}$ and $-\mathcal{R}$; a detailed comparison is shown in \autoref{sec:ablation}.}
We then obtain a new dataset $\POData{\rightarrow s} = \{(\textbf{x}_i\rightarrow (\textbf{t}^{w}_i, \textbf{t}^{l}_i), s) \mid (\textbf{x}_i, s_i)\in \Data\}$ for each style $s$.
Combining $\POData{\rightarrow s}$ for all styles, we finally obtain the PO dataset $\POData{}$.

\subsubsection{Reward Function}
\label{sec:reward}
To directly maximize the three objectives, F, MS, and, TSS, we use an aggregation of them as the reward function $\mathcal{R}$.
The most straightforward aggregation is to take the product of the three as in \citet{hallinan-etal-2023-steer}.
However, since the three objectives are independent, the probability of generating samples that have high scores in all three objectives is very low.
Our preliminary experiments show that samples with high total rewards can also have low single-objective scores, which naturally results in preference pairs in which the ``winning'' outputs have lower single-objective scores. We refer to these as \textit{reversed single-objective scores}.
When the percentage of reversed single-objective scores is high, we observe a degradation in the corresponding objective after PO.
To prevent the degradation in any objective, we propose to use a weighted product, which is given by
\begin{equation}
    \mathcal{R} = \text{TSS}^{\alpha}\cdot \text{MS}^{\beta}\cdot \text{F}^{\gamma}
\end{equation}
where $\alpha$, $\beta$, and $\gamma$ are temperatures for each objective.

We dynamically calculate $\alpha$, $\beta$, and $\gamma$ based on the number of reversed single-objective scores in the preference pairs for each iteration.
For convenience, we denote the number of reversed single-objective scores for each objective as $r_{\text{TSS}}$, $r_{\text{MS}}$, and $r_{\text{F}}$.\footnote{$r_{\text{TSS}}$, $r_{\text{MS}}$, and $r_{\text{F}}$ are functions of $\alpha$, $\beta$, and $\gamma$, so we recalculate $r$'s each time we change the value of $\alpha$, $\beta$, or $\gamma$.}
We first set $\beta=\gamma=1$ and set $\alpha$ to be the smallest positive integer such that $r_{\text{TSS}} < r_{\text{MS}}$ and $r_{\text{TSS}} < r_{\text{F}}$.
Then, we fix $\alpha$ and $\gamma$ and set $\beta$ to be the largest positive integer such that $r_{\text{MS}} > r_{\text{TSS}}$.
Finally, we fix $\alpha$ and $\beta$ and set $\gamma$ to be the largest positive integer such that $r_{\text{F}} > r_{\text{TSS}}$ and $r_{\text{F}} > r_{\text{MS}}$.
We set an upper bound $\tau_{\max}$ to $\alpha$, $\beta$, and $\gamma$ to prevent $\mathcal{R}$ from leaning too much to any objective.

\section{Experiments}

We evaluate STAMP on two text style transfer datasets in both in-domain and out-of-domain settings and compare STAMP with the state-of-the-art baseline approaches.
In this section, we detail the experimental setup and the model implementation.

\subsection{Datasets}
\label{sec:datasets}
We use two style transfer datasets in this work: (1) \textbf{Corpus of Diverse Styles (CDS)} \cite{krishna-etal-2020-reformulating}, which contains non-parallel texts in 11 different styles, such as Shakespeare and English Tweets, and (2) \textbf{Grammarly's Yahoo Answers Formality Corpus (GYAFC)} \cite{rao-tetreault-2018-dear}, which contains non-parallel formal and informal texts for training and a small number of parallel transfer pairs for tuning and test.
In this work, we only use non-parallel texts with style labels for training, validation, and test.

To reduce computational costs, we use a subset of each dataset.
Specifically, we sample 2000 texts per style for training, and 200 per style for validation. For CDS we sample 200 per style for test, while for GYAFC we sample 1000 per style. 
When constructing the end-to-end pseudo-parallel dataset $\TrfData$, for each target style, we sample 200 and 20 source texts from each of the other styles for training and validation, respectively.
In the in-domain testing, we transfer the test texts in each style to all other styles in the same dataset and calculate the total average scores and average scores grouped by the target style.
In the out-of-domain testing, we transfer all test texts in each dataset to all styles in the other dataset and calculate the same scores.
We elaborate on metric scores in \autoref{sec:metrics}.

Besides the style transfer datasets, we also use a paraphrase dataset, \textbf{ParaNMT} \cite{wieting-gimpel-2018-paranmt} to train the paraphraser used for pseudo-parallel data generation.
Specifically, we use the filtered version containing 75k paraphrase pairs in \citet{krishna-etal-2020-reformulating}.

\subsection{Reward Models}
\label{sec:reward_mdoels}
We have a reward model for each of the three objectives, TSS, MS, and F.
For convenience, we use the same notations to refer to the objective functions and the corresponding reward models in this paper.

\noindent\textbf{Target Style Strength (TSS)}\quad
We use a single style classifier, $f_{\text{cls}}$ with multiple binary sigmoid classification heads to calculate the TSS for each target style.
We train $f_{\text{cls}}$ from the pre-trained RoBERTa-large model \cite{liu2019roberta} on the same training and validation splits as discussed in \autoref{sec:datasets}.
We simply use the sigmoid outputs from the classification heads as the TSS scores which range from 0 to 1.

\noindent\textbf{Meaning Similarity (MS)}\quad
We assess the meaning similarity between the source text and the transferred text using the cosine similarity between the semantic embeddings of the two texts.
The semantic embeddings are calculated using SBERT\footnote{We use the variant with the best sentence embedding performance, which is all-mpnet-base-v2.} \cite{reimers-gurevych-2019-sentence}.
Technically, the cosine similarity of two embeddings ranges from -1 to 1, but negative cosine similarity is very rare in our experiments since we always the similarity between two paraphrases.
Following \citet{hallinan-etal-2023-steer}, we clip negative values to 0 to ensure that MS ranges from 0 to 1.

\noindent\textbf{Fluency (F)}\quad
To measure the fluency of a text, we use a text classifier\footnote{\url{https://huggingface.co/cointegrated/roberta-large-cola-krishna2020}} trained on the Corpus of Linguistic Acceptability (CoLA) \cite{warstadt-etal-2019-neural}.
The softmax score of the ``grammatical'' class is used as the F score which also ranges from 0 to 1.

\subsection{Baseline Approaches}
We compare STAMP with 4 strong baselines: GPT prompting \cite{reif-etal-2022-recipe}, STRAP \cite{krishna-etal-2020-reformulating}, STEER \cite{hallinan-etal-2023-steer}, and ASTRAPOP \cite{liu2024authorship}.

\noindent\textbf{GPT prompting} uses the zero- and few-shot capability of GPT-3.5-turbo to transfer texts to the target style given just the name of the style and 5 target style exemplars (5-shot) or no exemplars (zero-shot).

\noindent\textbf{STRAP} transfers a text by paraphrasing the text with a diverse paraphraser followed by an inverse paraphraser trained on pseudo-parallel transfer data generated by the diverse paraphraser.

\noindent\textbf{STEER} generates pseudo-parallel data using an expert-guided generation technique \cite{liu-etal-2021-dexperts}, and trains an end-to-end style transfer model on the generated data using a reinforcement learning algorithm \cite{lu2022quark}.

\noindent\textbf{ASTRAPOP} adopts the same paraphrase-and-inverse-paraphrase pipeline as STRAP but trains the inverse paraphraser using policy optimization or PO to directly maximize the target style strength, which achieves better performance on both low-resource and high-resource authorship style transfer. It does not use multi-iteration optimization, nor the overgeneration strategies we describe.

\subsection{Evaluation Metrics}

\subsubsection{Automatic Evaluation}
\label{sec:metrics}
We evaluate the approaches on the three objectives, TSS, MS, and F, using the same reward models introduced in \autoref{sec:reward_mdoels}.
To assess  overall performance, we use a single aggregate score $\text{Agg.} = \text{TSS}\cdot \text{MS}\cdot \text{F}$.
Note that the reward models described in \autoref{sec:reward_mdoels} calculate scores for single transfer pairs, while the final scores used for evaluation are averages over all transfer pairs in the test set.

\subsubsection{Human Evaluation}
\label{sec:human_metrics}
In addition to the automatic evaluation, we conduct a human evaluation to assess the model performance on the three style transfer objectives: $\text{TSS}_h$, $\text{MS}_h$, and $\text{F}_h$.\footnote{We use the subscript $h$ to distinguish human metrics from automatic metrics.}
For $\text{TSS}_h$, we show 5 exemplars for the style of the input text and 5 exemplars for the target style, and ask the annotator to select the style of the transferred text out of these two styles.
The sample gets a score of 1 if the target style is selected, and 0 otherwise.
For $\text{MS}_h$ and $\text{F}_h$, we ask whether the transferred text has a similar meaning to the input text and whether the transferred is fluent, respectively, and collect the answers using a three-level Likert scale ranging from 0 to 2.
See \autoref{sec:human_eval_inst} for the detailed instructions used in the human evaluation.

\begin{table*}[htb!]
\centering
\small
\setlength{\tabcolsep}{9pt}
\begin{tabular}{lcccclcccc}
\toprule
\multirow{2.5}{*}{Approach} & \multicolumn{4}{c}{CDS}                                           &  & \multicolumn{4}{c}{GYFAC}                                         \\
\cmidrule(rl){2-5}\cmidrule(rl){7-10}
                          & TSS            & MS             & F              & Agg.           &  & TSS            & MS             & F              & Agg.           \\
\midrule\midrule
GPT zero-shot             & 0.189$^\ddagger$          & 0.705$^\ddagger$          & 0.803$^\dagger$          & 0.104$^\ddagger$          &  & 0.672$^\ddagger$          & 0.788$^\ddagger$          & \textbf{0.968} & 0.489$^\ddagger$          \\
GPT 5-shot                & 0.199$^\ddagger$          & {\ul 0.735}$^\dagger$    & {\ul 0.805}$^\dagger$    & 0.112$^\ddagger$          &  & 0.667$^\ddagger$          & {\ul 0.800}$^\dagger$    & {\ul 0.965}    & 0.495$^\ddagger$          \\
STRAP                     & 0.382$^\ddagger$          & 0.626$^\ddagger$          & 0.759$^\ddagger$          & 0.158$^\ddagger$          &  & 0.618$^\ddagger$          & 0.735$^\ddagger$          & 0.913$^\ddagger$          & 0.409$^\ddagger$          \\
STEER                     & {\ul 0.654}$^\dagger$    & 0.672$^\ddagger$          & \textbf{0.905} & {\ul 0.395}$^\dagger$    &  & {\ul 0.951}    & 0.776$^\ddagger$          & 0.930$^\ddagger$          & {\ul 0.686}$^\dagger$    \\
ASTRAPOP                  & 0.542$^\ddagger$          & 0.600$^\ddagger$          & 0.755$^\ddagger$          & 0.221$^\ddagger$          &  & 0.783$^\ddagger$          & 0.734$^\ddagger$          & 0.924$^\ddagger$          & 0.525$^\ddagger$          \\
\midrule
STAMP                     & \textbf{0.746} & \textbf{0.801} & 0.801$^\dagger$          & \textbf{0.474} &  & \textbf{0.958} & \textbf{0.921} & 0.941$^\ddagger$          & \textbf{0.828} \\
\bottomrule
\end{tabular}
\caption{
    The automatic evaluation results on in-domain inputs on the CDS and the GYAFC datasets.
    The best and the 2\textsuperscript{nd} best scores in each column are shown in \textbf{bold} and {\ul underline}, respectively.
    ``$\dagger$'' and ``$\ddagger$'' indicate the score is significantly ($p<0.05$) worse than the best score and the top 2 scores in the same column, respectively, determined by resampling t-test.
}
\label{tab:in_domain}
\end{table*}

\subsection{Implementation Details}
\label{sec:implementation}
We implement all Seq2Seq models in STAMP, including the paraphraser and all transfer models, as decoder-only Seq2Seq models \cite{wolf2019transfertransfo} based on pre-trained LLaMA-2-7B \cite{touvron2023llama}.
The input and output are concatenated together with a separator token ``[SEP].''
For the unified transfer model $\SFTModel$, we prepend a style code for the target style (e.g., ``[SHAKESPEARE]'' and ``[FORMAL]'') to the input to control the output style.
We use CPO \cite{xu2024contrastive} in the multi-iteration PO stage.
We choose CPO instead of the most popular PO algorithm, DPO \cite{rafailov2023direct}, since CPO has been shown to be more efficient and effective \cite{xu2024contrastive, liu2024authorship}.
Also, compared to DPO, CPO has an additional negative log-likelihood term that is found to be significant for multi-iteration preference optimization \cite{pang2024iterative}.
We stop PO training at the iteration where the validation TSS starts to decrease and use the model from the previous iteration as the final model.
For fairness, all non-GPT baselines are also implemented based on LLaMA-2-7B and use the same paraphraser as STAMP.
We use gpt-3.5-turbo-0125 for all GPT-based approaches.
See \autoref{sec:more_implementation_details} for hyperparameters and GPT zero- and few-shot prompts.

\section{Results}

In this section, we present the quantitative experimental results.
A qualitative case study is in \autoref{sec:case_study}.
Because of the limited resources, we conduct all experiments for a single run and perform t-tests on the results.\footnote{See \autoref{ttest} for details.}

\subsection{Automatic Evaluation}

Automatic evaluation results on in-domain input are shown in \autoref{tab:in_domain}.
According to the aggregated score (Agg.), STAMP outperforms all baselines on the overall performance by a large margin on both datasets.
Looking at the per-objective scores, STAMP has the best target style strength (TSS) and meaning similarity (MS), but its fluency (F) is relatively lower, and this disadvantage is more obvious on the CDS dataset.
STEER has the best overall performance (Agg.) among the baselines on both datasets, while the overall performance of other baselines are mixed across the two datasets.
The results on the out-of-domain style transfer experiments are generally consistent with the in-domain results.
See \autoref{sec:out-of-domain} for details.

\begin{table}[h!]
\centering
\small
\begin{tabular}{lcccc}
\toprule
Approach   & TSS        & $\text{MS}_h/2$       & $\text{F}_h/2$         & $\text{Agg.}_{\sim h}$        \\
\midrule
GPT 5-shot & 0.16          & {\ul 0.75}    & {\ul 0.90}    & 0.11          \\
STEER      & {\ul 0.58}    & 0.62          & \textbf{0.92} & {\ul 0.33}    \\
STAMP      & \textbf{0.79} & \textbf{0.75} & 0.80          & \textbf{0.47} \\
\bottomrule
\end{tabular}
\caption{
    The human evaluation results on in-domain inputs on the CDS datasets.
    The best and the 2\textsuperscript{nd} best scores in each column are shown in \textbf{bold} and {\ul underline}, respectively.
}
\label{tab:human_eval_agg}
% \vspace{-0.2cm}
\end{table}

\begin{table*}[htb!]
\centering
\small
\setlength{\tabcolsep}{9pt}
\begin{tabular}{llllllllll}
\toprule
\multirow{2.5}{*}{Approach}        & \multicolumn{4}{c}{CDS}                                                                             &  & \multicolumn{4}{c}{GYFAC}                                                                           \\
\cmidrule(rl){2-5}\cmidrule(rl){7-10}
                                 & \multicolumn{1}{c}{TSS} & \multicolumn{1}{c}{MS} & \multicolumn{1}{c}{F} & \multicolumn{1}{c}{Agg.} &  & \multicolumn{1}{c}{TSS} & \multicolumn{1}{c}{MS} & \multicolumn{1}{c}{F} & \multicolumn{1}{c}{Agg.} \\
\midrule\midrule
STAMP                            & \textbf{0.746}          & 0.801$^\ddagger$                  & {\ul 0.801}$^\dagger$           & \textbf{0.474}           &  & 0.958$^\ddagger$                   & 0.921$^\dagger$                  & 0.941$^\dagger$                 & \textbf{0.828}           \\
\midrule
$\tau_{\mathcal{M}}=0.1$                     & 0.720$^\dagger$                   & 0.796$^\ddagger$                  & 0.800$^\dagger$                 & {\ul 0.454}$^\dagger$              &  & {\ul 0.965}             & 0.910$^\ddagger$                  & {\ul 0.943}$^\dagger$           & {\ul 0.826}              \\
$k_{\text{PO}} = 2$                        & {\ul 0.745}             & 0.688$^\ddagger$                  & \textbf{0.816}        & 0.411$^\ddagger$                    &  & \textbf{0.970}          & 0.878$^\ddagger$                  & \textbf{0.947}        & 0.804$^\ddagger$                    \\
Random $\textbf{t}^l$      & 0.640$^\ddagger$                   & \textbf{0.836}         & 0.780$^\ddagger$                 & 0.412$^\ddagger$                    &  & 0.950$^\ddagger$                   & {\ul 0.924}$^\dagger$            & 0.937                 & 0.822                    \\
High $\textbf{t}^l$ & 0.592$^\ddagger$                   & {\ul 0.826}$^\dagger$            & 0.796$^\dagger$                 & 0.384$^\ddagger$                    &  & 0.928$^\ddagger$                   & \textbf{0.936}         & 0.932$^\ddagger$                 & 0.810$^\ddagger$  \\
\bottomrule
\end{tabular}
\caption{
    Hope-and-fear sampling ablations, evaluated automatically on in-domain inputs on the CDS and the GYAFC datasets.
    The best and the 2\textsuperscript{nd} best scores in each column are shown in \textbf{bold} and {\ul underline}, respectively.
    ``$\dagger$'' and ``$\ddagger$'' indicate the score is significantly ($p<0.05$) worse than the best score and the top 2 scores in the same column, respectively, determined by resampling t-test.
}
\label{tab:ablation}
% \vspace{-0.2cm}
\end{table*}

\subsection{Human Evaluation}
We conduct a human evaluation on the CDS dataset for STAMP, the best-performing baseline (STEER), and the best GPT-prompting baseline (GPT 5-shot).
We randomly choose 5 samples from each of the 11 target styles for each of the three models, which yields 165 samples in total, and collect up to three annotations for each sample.
Seven volunteer NLP experts are recruited for annotation.
We perform an independent sample t-test on the annotation results and find statistically significant differences in $\text{MS}_h$ and $\text{F}_h$ but not in $\text{TSS}_h$,\footnote{See \autoref{sec:more_human_eval} for the raw human evaluation scores and the result of the t-test.} which is in line with our expectation since the style classification has been found to be hard for untrained humans \cite{krishna-etal-2020-reformulating, hallinan-etal-2023-steer}.
Therefore, following \citet{krishna-etal-2020-reformulating} and \citet{hallinan-etal-2023-steer}, we calculate the quasi aggregated score $\text{Agg.}_{\sim h}$ using TSS,\footnote{which is calculated from the human study samples using the automatic TSS metric.} $\text{MS}_h$, and $\text{F}_h$.
Formally, $\text{Agg.}_{\sim h} = \text{TSS}\cdot\frac{\text{MS}_h}{2}\cdot\frac{\text{F}_h}{2}$, where we divide $\text{MS}_h$ and $\text{F}_h$ by 2 to scale them to the [0, 1] range so that $\text{Agg.}_{\sim h}$ also ranges from 0 to 1.
As shown in \autoref{tab:human_eval_agg}, STAMP has the best meaning similarity ($\text{MS}_h$) and overall performance ($\text{Agg.}_{\sim h}$), but its fluency is worse than STEER and GPT 5-shot transfer, which is consistent with the automatic evaluation results.

\subsection{Ablation Studies}
\label{sec:ablation}

In this section, we demonstrate the effects of our four main contributions in STAMP: multi-iteration PO, hope-and-fear sampling, weighted reward aggregation, and end-to-end pseudo-parallel data generation.

\begin{figure}[h!]
    \centering
    \includegraphics[width=0.75\columnwidth]{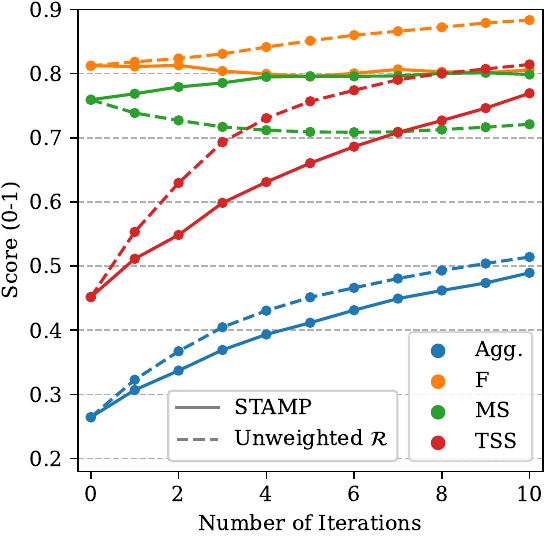}
    \caption{The value of iterative CPO on performance in STAMP and STAMP with unweighted $\mathcal{R}$, shown on the CDS dataset (test split). Iteration 0 refers to the SFT model before PO.}
    \label{fig:cds_multi-iter}
    % \vspace{-0.2cm}
\end{figure}

\noindent\textbf{Multi-iteration PO \& Weighted $\mathcal{R}$}\quad 
We show the performance evolution of STAMP and STAMP with unweighted $\mathcal{R}$ over the multi-iteration PO training in \autoref{fig:cds_multi-iter}.
In general, the overall performance (Agg.)\ of both models keeps increasing over the iterations, which indicates the effectiveness of multi-iteration optimization.
STAMP with unweighted $\mathcal{R}$ performs slightly better than STAMP, but it has a severe degradation in meaning similarity (MS), and the scores in the three objectives have a substantial difference after training.
In contrast, with the weighted reward aggregation, STAMP shows a higher stability in all scores.
Only fluency (F) exhibits a slight decrease, and scores in all three objectives converge to a similar value at the end of the training. 

\noindent\textbf{Hope-and-fear Sampling}\quad
The results of hope-and-fear sampling ablation are shown in \autoref{tab:ablation}.
As mentioned in \autoref{sec:reward}, we do not use the model score term in hope-and-fear sampling for preference pair construction since it does not improve the performance, which can be observed from the ``$\tau_{\mathcal{M}}=0.1$'' row in \autoref{tab:ablation}.
The last three rows in \autoref{tab:ablation} show that both dropping over-generation ($k_{\text{PO}}=2$) and using a random other sample (Random $t^l$) or the sample with the second highest reward (High $\textbf{t}^l$) as the ``losing'' sample undermine the overall performance of STAMP.

\noindent\textbf{Pseudo-parallel Data Generation}\quad
We demonstrate the superiority of our two-step end-to-end pseudo-parallel data generation method by comparing the STAMP SFT model, $\SFTModel$, with the best-performing baseline SFT style transfer model, STRAP.
The overall performance (Agg.) of the two models is shown in \autoref{tab:ablation_sft}.
With our method, the overall performance of $\SFTModel$ is much higher than STRAP on both datasets, which provides a better starting point for PO.

\begin{table}[htb!]
\centering
\small
\begin{tabular}{l@{\hspace{30pt}}cc}
\toprule
          & CDS   & GYAFC \\
\midrule
STRAP     & 0.158 & 0.409 \\
$f_{SFT}$ & \textbf{0.264} & \textbf{0.657} \\
\bottomrule
\end{tabular}
\caption{
    The overall performance (Agg.) of STRAP and the STAMP SFT model ($f_{SFT}$) on CDS and GYAFC.
    The best score in each column is shown in \textbf{bold}.
}
\label{tab:ablation_sft}
\vspace{-0.1cm}
\end{table}
\section{Related Work}

\noindent\textbf{Text Style Transfer}\quad
% \subsection{Text Style Transfer}
Due to the lack of parallel style transfer data, only a limited number of studies address this task as a supervised or semi-supervised Seq2Seq task, which requires a certain amount of parallel data for training and/or tuning \cite{zhu-etal-2010-monolingual, rao-tetreault-2018-dear, wang-etal-2019-harnessing, shang-etal-2019-semi, xu2019formality, zhang-etal-2020-parallel, kim-etal-2022-improving, raheja-etal-2023-coedit}.
Although these approaches work well when parallel data is available, none generalize well to styles with no parallel data. 
As a result, most works in this area focus on unsupervised approaches that require only non-parallel data or even no data.
These works mainly approach the task via latent representation disentanglement and manipulation \cite{lample2018multipleattribute, Liu2019RevisionIC, john-etal-2019-disentangled, jin-etal-2020-hooks}, style-related pattern editing \cite{madaan-etal-2020-politeness, malmi-etal-2020-unsupervised, reid-zhong-2021-lewis, luo-etal-2023-prompt}, pseudo-parallel transfer data construction \cite{krishna-etal-2020-reformulating, riley-etal-2021-textsettr}, policy optimization \cite{gong-etal-2019-reinforcement, liu-etal-2021-learning, deng-etal-2022-rlprompt, hallinan-etal-2023-steer, liu2024authorship}, and LLM zero- or few-shot prompting \cite{reif-etal-2022-recipe, suzgun-etal-2022-prompt, patel2023lowresource}.

Among these approaches, two of the policy optimization based approaches, STEER \cite{hallinan-etal-2023-steer} and ASTRAPOP \cite{liu2024authorship} achieve the best performance on text style transfer and authorship style transfer, respectively.
Their high-level training frameworks both combine pseudo-parallel data generation and policy optimization, but their specific approaches differ.
For pseudo-parallel data generation, STEER uses a paraphraser guided by an expert and an anti-expert, while ASTRAPOP simply paraphrases the texts in the target style and uses these paraphrase-to-target transfer pairs.
For policy optimization, STEER uses an RL algorithm, Quark, while ASTRAPOP tries three options: one RL algorithm, PPO \cite{schulman2017proximal}, and two PO algorithms, DPO \cite{rafailov2023direct} and CPO \cite{xu2024contrastive}.
Our framework shares the same high-level procedure with STEER and ASTRAPOP, but we design a new pseudo-parallel data generation method and also enhance the PO stage with multi-iteration training, weighted reward aggregation, and hope-and-fear preference pair construction, These enhancements dramatically improve the performance of STAMP over STEER and ASTRAPOP.

\noindent\textbf{Preference Optimization}\quad
% \subsection{Preference Optimization}
PO \cite{rafailov2023direct, song2024preference, xu2024contrastive} is a class of RL-free policy optimization algorithms which has been broadly applied to train generative language models on direct task objectives instead of the language modeling loss and is closely related to (pre-neural) machine translation objective `tuning' \cite{och-2003-minimum,chiang-etal-2008-online,hopkins-may-2011-tuning}.
\citet{rafailov2023direct} show that PO is more stable and efficient than traditional RL-based algorithms on sentiment generation and text summarization \cite{rafailov2023direct}.
It has also been successfully applied to many other NLP tasks, such as training helpful and harmless assistants \cite{song2024preference}, machine translation \cite{xu2024contrastive}, and authorship style transfer \cite{liu2024authorship}.
Later works \cite{xiong2023iterative, xu2024things, yuan2024selfrewarding, chen2024selfplay, pang2024iterative, song2024trial} extend the offline PO algorithms by performing the optimization for multiple iterations and further improve the performance of the models.
In this work, we adopt the multi-iteration PO for STAMP and enhance it with weighted reward aggregation and hope-and-fear preference pair construction, which improve the effectiveness of multi-iteration PO training.% over vanilla iterative PO.

\section{Conclusion}

We present STAMP, a multi-iteration preference optimization training framework for text style transfer, in which an end-to-end pseudo-parallel data generation pipeline provides a strong reference model, a preference pair construction strategy improves the effectiveness of PO training, and weighted reward aggregation ensures  balance across multiple objectives over multi-iteration training.
We evaluate STAMP on two commonly used text style transfer datasets;
demonstrating superior performance over all state-of-the-art style transfer approaches.

\section*{Limitations}

Although achieving the state-of-the-art performance on two text style transfer datasets, STAMP has two main limitations.
First, we observe repetitions and hallucinations in some transferred texts.
The potential reason is that PO training increases the peakiness of the model, which means the probability of generating the tokens that are frequent in the target style increases disproportionately \cite{Choshen2020On, kiegeland-kreutzer-2021-revisiting}.
The occurrence of repetitions and hallucinations also indicates that our reward model cannot fully capture all aspects of the desired objectives.
Two possible solutions are developing PO algorithms that are less vulnerable to the increased peakiness and developing better reward models.
These are two promising directions for future studies but are out of the scope of the current work which focuses on the multi-iteration extension of existing preference optimization algorithms and the strategies for preference pair construction.

Second, as discussed in \autoref{sec:ablation}, the weighted reward aggregation method is effective on the CDS dataset but is not very useful on the GYAFC dataset because formality transfer is a relatively easier task, and it is more likely to generate high-quality samples with balanced single-objective scores.
It could be useful to add a control mechanism to determine when using the weighted aggregation is beneficial to prevent overbalanced single-objective scores on easy tasks.
\section*{Ethical Considerations}

As a general text style transfer framework, STAMP can transfer texts to any target style given an adequate amount of non-parallel data, which means it can potentially be used to generate unethical texts such as transferring normal texts into an offensive or profane style.
Moreover, although STAMP is not specifically designed for authorship transfer, it can still serve that purpose by transferring the texts into the style of a particular author, which can be unethical if used without authorization. However, privatization of an author's style can also be used to enable oppressed people to communicate freely without the fear of recrimination. In any case, as we and others show, the state of the art of style transfer is not yet advanced for either privacy or mimicry to be a significant concern in a deployed system. Our work is strictly intended for research and personal use on public or authorized data.

Some texts in the datasets used in this work (though collected and released elsewhere) contain words or ideas that may cause harm to others. We do not generally filter out those texts, so that we may maximally preserve the characteristics of different styles. However, for human studies, we remove all texts with personal identifiable information (PII) to ensure privacy and remove texts that contain profane language to minimize harm to human subjects. We exclude these texts instead of masking out PII or profane tokens, since masks may influence annotators' judgments regarding meaning similarity and fluency. The protocols of our human studies have been approved by an institutional review board.

\section*{Acknowledgments}

% chiron = HR00112490374.
% hiatus = 2022-22072200006
This research is supported in part by the Office of the Director of National Intelligence (ODNI), Intelligence Advanced Research Projects Activity (IARPA), via the HIATUS Program contract \#2022-22072200006, and in part by the Defense Advanced Research Projects Agency (DARPA) under Agreement No. HR00112490374. The views and conclusions contained herein are those of the authors and should not be interpreted as necessarily representing the official policies, either expressed or implied, of ODNI, IARPA, DARPA, or the U.S. Government. The U.S. Government is authorized to reproduce and distribute reprints for governmental purposes notwithstanding any copyright annotation therein.  

\bibliography{anthology,custom}

\appendix

\begin{table*}[htb!]
\centering
\small
\setlength{\tabcolsep}{9pt}
\begin{tabular}{lcccclcccc}
\toprule
\multirow{2.5}{*}{Approach} & \multicolumn{4}{c}{CDS}                                           &  & \multicolumn{4}{c}{GYFAC}                                         \\
\cmidrule(rl){2-5}\cmidrule(rl){7-10}
                          & TSS            & MS             & F              & Agg.           &  & TSS            & MS             & F              & Agg.           \\
\midrule\midrule
GPT zero-shot             & 0.246$^\ddagger$          & 0.657$^\ddagger$          & 0.855$^\ddagger$          & 0.138$^\ddagger$          &  & 0.672$^\ddagger$          & 0.752$^\dagger$          & \textbf{0.909} & 0.455$^\ddagger$          \\
GPT 5-shot                & 0.289$^\ddagger$          & {\ul 0.708}$^\dagger$    & 0.868$^\dagger$          & 0.175$^\ddagger$          &  & 0.722$^\ddagger$          & {\ul 0.752}$^\dagger$    & {\ul 0.902}    & 0.486$^\ddagger$          \\
STRAP                     & 0.426$^\ddagger$          & 0.629$^\ddagger$          & 0.810$^\ddagger$          & 0.194$^\ddagger$          &  & 0.692$^\ddagger$          & 0.689$^\ddagger$          & 0.852$^\ddagger$          & 0.402$^\ddagger$          \\
STEER                     & {\ul 0.654}$^\dagger$    & 0.706$^\dagger$          & \textbf{0.927} & {\ul 0.426}$^\dagger$    &  & {\ul 0.850}$^\dagger$    & 0.734$^\ddagger$          & 0.875          & {\ul 0.544}$^\dagger$    \\
ASTRAPOP                  & 0.579$^\ddagger$          & 0.606$^\ddagger$          & 0.808$^\ddagger$          & 0.259$^\ddagger$          &  & 0.816$^\dagger$          & 0.685$^\ddagger$          & 0.863$^\ddagger$          & 0.479$^\ddagger$          \\
\midrule
STAMP                     & \textbf{0.787} & \textbf{0.816} & {\ul 0.877}$^\dagger$    & \textbf{0.562} &  & \textbf{0.964} & \textbf{0.864} & 0.827$^\ddagger$          & \textbf{0.687} \\
\bottomrule
\end{tabular}
\caption{
    The automatic evaluation results on out-of-domain inputs on the CDS and the GYAFC datasets.
    The best and the 2\textsuperscript{nd} best scores in each column are shown in \textbf{bold} and {\ul underline}, respectively.
    ``$\dagger$'' and ``$\ddagger$'' indicate the score is significantly ($p<0.05$) worse than the best score and the top 2 scores in the same column, respectively, determined by resampling t-test.
}
\label{tab:out_of_domain}
\end{table*}

\section{More Experimental Results}
\label{sec:more_experimental_resutls}

\subsection{Out-of-domain Style Transfer}
\label{sec:out-of-domain}

\autoref{tab:out_of_domain} shows  automatic evaluation results of the `out-of-domain' style transfer experiments, in which we transfer the texts in each dataset to the styles in the other dataset, in order to determine whether our results hold up when transferring between styles of different provenance.
They do; the out-of-domain results are generally consistent with the in-domain results.
The best model in each column in \autoref{tab:out_of_domain} is the same as \autoref{tab:in_domain}, which is also true for the second best model in most columns.
Also, STAMP still has the best TSS, MS, and aggregated score (Agg.) among all approaches, and STEER still has the best overall performance (Agg.) among the baselines. 

\subsection{More Human Evaluation Results}
\label{sec:more_human_eval}

\begin{table}[htb!]
\centering
\begin{tabular}{l@{\hspace{30pt}}ccc}
\toprule
Approach   & $\text{TSS}_h$ & $\text{MS}_h$ & $\text{F}_h$ \\
\midrule
GPT 5-shot & 0.59          & {\ul 1.48}    & {\ul 1.79}    \\
STEER      & \textbf{0.69} & 1.24$^\ddagger$          & \textbf{1.84} \\
STAMP      & {\ul 0.64}    & \textbf{1.48} & 1.57$^\ddagger$       \\
\bottomrule
\end{tabular}
\caption{
    Raw human evaluation scores on in-domain inputs on the CDS datasets.
    The best and  2\textsuperscript{nd} best scores in each column are shown in \textbf{bold} and {\ul underline}, respectively.
    ``$\ddagger$'' indicates a statistically significant difference ($p<0.05$) between the top two models determined by independent sample t-test.
    No significant difference is found in any other model pairs.
}
\label{tab:human_eval_raw}
\end{table}

The raw scores from the human evaluation and the result of the t-test are shown in \autoref{tab:human_eval_raw}.
No significant difference is found between any model pairs in $\text{TSS}_h$\footnote{which is expected since style classification is difficult for human annotators \cite{krishna-etal-2020-reformulating, hallinan-etal-2023-steer}.}, but $\text{MS}_h$ and $\text{F}_h$ are generally consistent with the automatic evaluation results.
Specifically, STAMP and GPT 5-shot transfer are significantly better than STEER in meaning similarity (MS), and STEER and GPT 5-shot transfer are significantly better than STAMP in fluency (F).

\subsection{Case Study}
\label{sec:case_study}

We show an example from the CDS test set in \autoref{tab:case_study} as a case study.
In this example, we transfer a text in the style of music lyrics into the style of Shakespeare using STAMP and all baseline approaches.
STAMP maximally preserves the meaning of the original sentence and accurately reflects the target style using the words "'tis" and "o'er".
Other approaches either fail to generate strong target style indicators or change the meaning of the original sentence to some extent.

\begin{table*}[]
\centering
\begin{tabular}{ll}
\toprule
Original      & I'm not sorry that it's over                                                   \\
\midrule
GPT zero-shot & I doth not lament the end that hath befallen us.                               \\
GPT 5-shot    & I doth not lament the ending of this tale.                                     \\
STRAP         & I am not sorry That he is gone.                                                \\
ASTRAPOP      & Now is the winter of our discontent Made glorious summer by this sun of York. \\
STEER         & I do not regret that it is done.                                               \\
STAMP         & I am not sorry That ’tis o’er.                                                 \\
\bottomrule
\end{tabular}
\caption{
    A style transfer example from the style of music lyrics to the style of Shakespeare.
}
\label{tab:case_study}
\end{table*}
\begin{table}[]
\centering
\begin{tabular}{lcccc}
\toprule
Parameter                & $f_{cls}$ & $f_{para}$ & $f_{p\rightarrow t}$ & $f_{s\rightarrow t}$ \\
\midrule
learning rate   & 5e-5 & 5e-5    & 5e-5        & 5e-5        \\
batch size      & 32 & 32      & 8           & 16          \\
\# epochs       & 6 & 10      & 6           & 12          \\
\bottomrule
\end{tabular}
\caption{
    Training hyperparameters for all supervised fine-tuned models.
}
\label{tab:hyperparameters_sft}
\end{table}

\begin{table}[]
\centering
\begin{tabular}{l@{\hspace{80pt}}c}
\toprule
Parameter                & $f_{PO}$ \\
\midrule
learning rate   & 2e-6  \\
$\beta$           & 0.1  \\
batch size      & 32    \\
\# epochs       & 16    \\
$k_{\text{PO}}$ & 10    \\
$N_{\text{iter}}$         & 10    \\
\bottomrule
\end{tabular}
\caption{
    Training hyperparameters for iterative preference optimization.
}
\label{tab:hyperparameters_po}
\end{table}

\begin{table}[]
\centering
\begin{tabular}{l@{\hspace{40pt}}r}
\toprule
Parameter      &                  \\
\midrule
target modules & q\_proj, v\_proj \\
rank           & 16               \\
$\alpha$          & 32               \\
dropout        & 0.05             \\
\bottomrule
\end{tabular}
\caption{
    LoRA Hyperparameters.
}
\label{tab:hyperparameters_lora}
\end{table}

\begin{table}[]
\centering
\begin{tabular}{lccc}
\toprule
Parameter            & $D_{p\rightarrow t}$ & $D_{s\rightarrow t}$ & $D_{PO}$ \\
\midrule
top p       & 1.0                             & 1.0                             & 1.0                       \\
temperature & 0.5                             & 0.7                             & 1.0                       \\
$k_{\text{para/sft/po}}$     & 20                              & 90                               & 10                         \\
$\tau_{text{MS/max}}$         & -       & 8           & 6   \\
\bottomrule
\end{tabular}
\caption{
    Generation hyperparameters for dataset construction.
}
\label{tab:hyperparameters_data_gen}
\end{table}

\begin{table}[]
\centering
\begin{tabular}{l@{\hspace{70pt}}c}
\toprule
Parameter            & Evaluation \\
\midrule
top p       & 1.0       \\
temperature & 0.7       \\
\bottomrule
\end{tabular}
\caption{
    Generation hyperparameters for dataset evaluation.
}
\label{tab:hyperparameters_eval}
\end{table}

\begin{table*}[]
\centering
\begin{tabular}{ll}
\toprule
Zero-shot & \begin{tabular}[c]{@{}l@{}}\texttt{Rewrite the following sentence into the style of {[}target style{]}.}\\ \texttt{Original Sentence: {[}input text{]}}\\ \texttt{Rewritten Sentence: }\end{tabular} \\
\midrule
5-shot    & \begin{tabular}[c]{@{}l@{}}\texttt{Here are some examples of sentences in the style of {[}target style{]}:}\\ \texttt{{[}example 1{]}}\\ \texttt{......}\\ \texttt{{[}example 5{]}}\\ \texttt{Rewrite the following sentence into the style of {[}target style{]}.}\\ \texttt{Original Sentence: {[}input text{]}}\\ \texttt{Rewritten Sentence: }\end{tabular}  \\
\bottomrule
\end{tabular}
\caption{
    GPT zero- and 5-shot prompts for style transfer on CDS.
}
\label{tab:gpt_prompt_cds}
\end{table*}

\begin{table*}[]
\centering
\begin{tabular}{ll}
\toprule
Zero-shot & \begin{tabular}[c]{@{}l@{}}\texttt{Rewrite the following sentence in a(n) (in)formal style.}\\ \texttt{Original Sentence: {[}input text{]}}\\ \texttt{Rewritten Sentence: }\end{tabular} \\
\midrule
5-shot    & \begin{tabular}[c]{@{}l@{}}\texttt{Here are some examples of sentences in a(n) (in)formal style:}\\ \texttt{{[}example 1{]}}\\ \texttt{......}\\ \texttt{{[}example 5{]}}\\ \texttt{Rewrite the following sentence in a(n) (in)formal style.}\\ \texttt{Original Sentence: {[}input text{]}}\\ \texttt{Rewritten Sentence: }\end{tabular}  \\
\bottomrule
\end{tabular}
\caption{
    GPT zero- and 5-shot prompts for style transfer on GYAFC.
}
\label{tab:gpt_prompt_gyafc}
\end{table*}

\begin{table*}[]
\centering
\begin{tabular}{lccccccccc}
\toprule
            & ParaNMT & \multicolumn{4}{c}{CDS}           & \multicolumn{4}{c}{GYAFC}         \\
\cmidrule(rl){2-2}\cmidrule(rl){3-6}\cmidrule(rl){7-10}
            & $f_{para}$ & $f_{cls}$ & $f_{p\rightarrow t}$ & $f_{s\rightarrow t}$ & $f_{PO}$ & $f_{cls}$ & $f_{p\rightarrow t}$ & $f_{s\rightarrow t}$ & $f_{PO}$ \\
\midrule
\# GPUs (A40s)       & $\times$2  & $\times$2 & $\times$2       & $\times$2       & $\times$4 & $\times$2 & $\times$2       & $\times$2       & $\times$2 \\
Times (hrs) & 3.4    & 0.4  & 1.1         & 1.0         & 35.2  & 0.1 & 0.2         & 0.2         & 7.4   \\
\bottomrule
\end{tabular}
\caption{
    Training hardware and runtime for each component in STAMP on CDS and GYAFC.
}
\label{tab:gpu_and_runtime}
\end{table*}

\begin{table*}[]
\centering
\begin{tabular}{llc}
\toprule
Type                     & Name                                              & License                        \\
\midrule
\multirow{2}{*}{Dataset} & \href{https://github.com/martiansideofthemoon/style-transfer-paraphrase}{CDS: Corpus of Diverse Styles}                   & MIT                            \\
                         & \href{https://github.com/raosudha89/GYAFC-corpus}{GYAFC: Grammarly’s Yahoo Answers Formality Corpus} & Custom (research-only)                             \\
\midrule
\multirow{5}{*}{Model}   & \href{https://huggingface.co/meta-llama/Llama-2-7b-hf}{LLaMA-2-7B} (6.7B)                                 & Meta                           \\
                         & \href{https://platform.openai.com/docs/models/gpt-3-5-turbo}{GPT-3.5-turbo-0125} (-)                            & MIT                            \\
                         & \href{https://huggingface.co/FacebookAI/roberta-large}{RoBERTa-large} (355M)                              & MIT                            \\
                         & \href{https://huggingface.co/cointegrated/roberta-large-cola-krishna2020}{RoBERTa-large CoLA Classifier} (355M)                            & MIT                            \\
                         & \href{https://huggingface.co/sentence-transformers/all-mpnet-base-v2}{SBERT all-mpnet-base-v2} (109M)                    & Apache-2.0 \\
\midrule
\multirow{4}{*}{Library} & \href{https://github.com/huggingface/transformers}{Transformers}                                      & Apache-2.0                     \\
                         & \href{https://github.com/huggingface/peft}{PEFT}                                              & Apache-2.0 \\
                         & \href{https://github.com/huggingface/trl}{TRL}                                               & Apache-2.0 \\
                         & \href{https://github.com/UKPLab/sentence-transformers}{Sentence Transformers}                             & Apache-2.0 \\
\bottomrule
\end{tabular}
\caption{
    Datasets, models, and software libraries used in this work.
    The number of parameters of each model is indicated in the parentheses next to the model name.
}
\label{tab:artifacts}
\end{table*}

\begin{table*}[htb!]
\centering
\begin{tabularx}{\textwidth}{llX}
\toprule
$\text{TSS}_h$                 & Question         & Based on the examples above, what is the style of the following text?                                                                     \\
\midrule
\multirow{6}{*}{$\text{MS}_h$} & Similar          & Most of the meaning (75\% or more) of the two passages is the same.                                                                       \\
\cmidrule(rl){2-3}
                    & Somewhat Similar & Large portions (50-75\%) of the passages are the same, but there are significant sections that differ or are present in only one passage. \\
\cmidrule(rl){2-3}
                    & Not Similar      & Only small portions (less than 50\%) of the passages are the same.                                                                        \\
\cmidrule(rl){2-3}
                    & Question         & How similar are the following two texts?                                                                                                  \\
\midrule
\multirow{7.5}{*}{$\text{F}_h$}  & Fluent           & Very clear, grammatical english (need not be formal); the meaning of the sentence is well understood. A small number of errors are ok.    \\
\cmidrule(rl){2-3}
                    & Somewhat Fluent  & There are grammatical errors, possibly numerous, but the meaning can be understood.                                                       \\
\cmidrule(rl){2-3}
                    & Not Fluent       & The grammatical errors make it very difficult to understand the meaning.                                                                  \\
\cmidrule(rl){2-3}
                    & Question         & How fluent is the following text?           \\
\bottomrule
\end{tabularx}
\caption{
    Instructions used in the human evaluation.
}
\label{tab:human_eval_inst}
\end{table*}

\section{More Implementation Details}
\label{sec:more_implementation_details}

\subsection{Statistical Significance Test}
\label{ttest}

We conduct a resampling paired t-test for the automatic evaluation results and an independent t-test for the human evaluation results.
For the resampling paired t-test, we randomly select 10 subsets of 100 samples from the test set and perform a paired t-test on the mean scores of the subsets between each pair of models.
For the independent t-test, we use all available samples from the human study without resampling.

\subsection{Hyperparameters}

We sample same-sized training and validation subsets for CDS and GYAFC, and use the same hyperparameters to train STAMP on the two datasets to reduce the cost for more hyperparameter searching.
We list all hyperparameters for STAMP in \autoref{tab:hyperparameters_sft}, \autoref{tab:hyperparameters_po}, \autoref{tab:hyperparameters_lora}, \autoref{tab:hyperparameters_data_gen}, and \autoref{tab:hyperparameters_eval}.

\subsection{GPT prompt templates}
We elaborate on the prompts used for GPT zero- and 5-shot style transfer on CDS and GYAFC in \autoref{tab:gpt_prompt_cds} and \autoref{tab:gpt_prompt_gyafc}, respectively.

\subsection{Hardware and Runtime}
We train all components of STAMP using Nvidia A40-48GB GPUs. The number of GPUs and time used to train each model on each dataset are shown in \autoref{tab:gpu_and_runtime}.

\subsection{Human Evaluation Instructions}
\label{sec:human_eval_inst}
The instructions used in the human evaluation for all three objectives are shown in \autoref{tab:human_eval_inst} including the questions asked and the detailed explanation for each level in the Likert scale.

\section{Scientific Artifacts}
\label{sec:artifacts}

\subsection{Use of Existing Artifacts}
The existing artifacts used in this work and their licenses are listed in \autoref{tab:artifacts}.
Our use of the existing artifacts is consistent with their intended use specificed by their licenses.

\subsection{Created Artifacts}
We create a new text style transfer training framework, STAMP, and release the code under the MIT license.
Considering ethical implications, STAMP is only intended for research purposes, which is compatible with the original access conditions of all existing artifacts used in STAMP.

\end{document}